%% file: main.tex
\documentclass{article}

\usepackage[preprint]{corl_2026}      
\usepackage{enumitem}
\usepackage{amsmath, amssymb}
\usepackage{graphicx}
\usepackage{booktabs}
\usepackage{xspace}
\usepackage{tikz}
\usetikzlibrary{positioning, fit, calc, decorations.pathreplacing, arrows.meta, shapes.geometric}

\newcommand{\method}{STAIR\xspace}   
\newcommand{\fdyn}{\mathbf{f}^{\text{dyn}}}

\usepackage{wrapfig}
\usepackage[table]{xcolor}
\usepackage{multirow}
\title{%
  \texorpdfstring{%
    \includegraphics[height=1.25em]{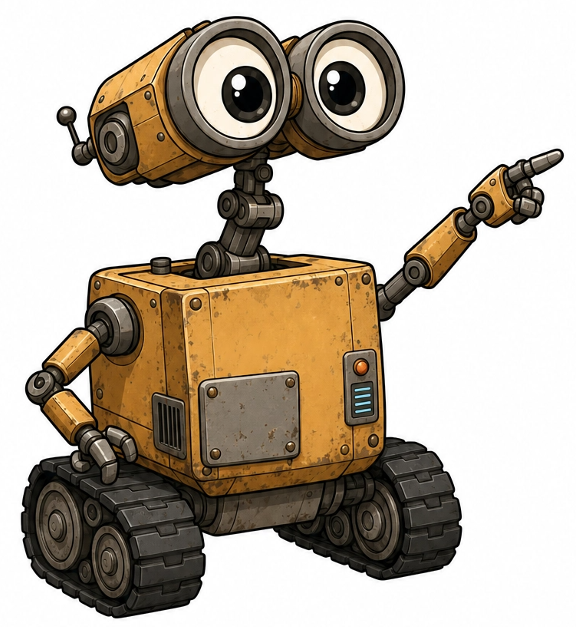}%
    Perfect Demo Makes Poor Teacher: Learning Robust Alignment from Critical Motion Segments%
  }%
}

\author{
 Mingyu Liu$^{1,2}$\thanks{Equal contribution. $^{\dagger}$Corresponding author.}\quad 
 Zeju Li$^{1*}$\quad
 Jiuhe Shu$^{1}$\quad \\
 \textbf{Hanqing Wang}$^{3}$\quad
 \textbf{Yuhao Chao}$^{2,4}$\quad
 \textbf{Hao Chen}$^{1}$\quad 
 \textbf{Chunhua Shen}$^{1\dagger}$\quad\\
 $^{1}$Zhejiang University \quad
 $^{2}$Shanghai Innovation Institute \quad \\
 $^{3}$Hong Kong University of Science and Technology (GZ) \quad
 $^{4}$Nanjing University \quad
 \\
  \small{\href{https://github.com/aim-uofa/STAIR}{{\color{orange} \textbf{github.com/aim-uofa/STAIR}}}}
  \vspace{-1.5em}
}

\begin{document}
\maketitle

\input{sections/0_Abstract}
\input{sections/1_Intro}
\input{sections/2_Related_Works}
\input{sections/3_Method}
\input{sections/4_Exe}
\input{sections/5_Conclusion}
\input{sections/6_limitation}

\clearpage

\bibliography{main}


\end{document}

%% file: sections/0_Abstract.tex
\begin{abstract}
Expert demonstrations are widely assumed to be the gold standard for robot imitation learning. Yet for fine-grained manipulation such as insertion, stacking, and alignment, we uncover a counterintuitive failure mode: fluent demonstrations can be poor teachers. A skilled teleoperator compresses the decisive moments of alignment and recovery into a brief temporal window, leaving the policy flooded with redundant free-space motion and starved of supervision exactly where precision determines success. We address this bottleneck at two levels. At the data level, slowing down near alignment and resampling critical segments both help, yet the gain comes mainly from broadening the coverage of recovery states the policy must learn, not from reweighting frames it already has. Such data-side fixes, however, leave the policy's per-frame view untouched: a single image still maps directly to an action, and the local motion that governs correction stays implicit. We therefore turn to the representation level and introduce \method (\textbf{S}patio-\textbf{T}emporal feature \textbf{A}s an \textbf{I}nterface for \textbf{R}obot learning), a compact dynamic feature that bridges the vision-language model and the action expert, distilling the short-horizon motion already recorded in each trajectory into dense, motion-aware supervision. Trained on fluent data alone, \method recovers most of the deliberate-demonstration gain ($50.0$ to $62.2\%$ overall, approaching the $64.4\%$ of deliberate demonstrations). These results call for a more pedagogical view of robot data, optimized for machine learnability rather than human efficiency alone.
\end{abstract}

\keywords{Robot Manipulation, Representation Learning, Data Curation}

%% file: sections/1_Intro.tex
\section{Introduction}
\label{sec:intro}

\emph{``In the beginner's mind there are many possibilities, but in the expert's there are few.''}\\
\hspace*{\fill}--- Shunryu Suzuki, \emph{Beginner's Mind}

\textbf{Robots, in this sense, are still beginners.} A skilled teleoperator can make fine manipulation look effortless. In tasks such as block stacking or tight insertion, the motion looks smooth from start to finish: a fast approach, a brief final adjustment, a clean completion. Such demonstrations seem ideal for imitation learning~\citep{ravichandar2020recent}. Yet in practice we observe the opposite. Policies trained on these fluent demonstrations master the approach but fail in the last few centimeters, where a small alignment error turns into collision or unrecoverable drift. This exposes a simple tension: \emph{a perfect demonstration can make a poor teacher.}

\begin{figure}[h]
\begin{center}
\includegraphics[width=1\linewidth]{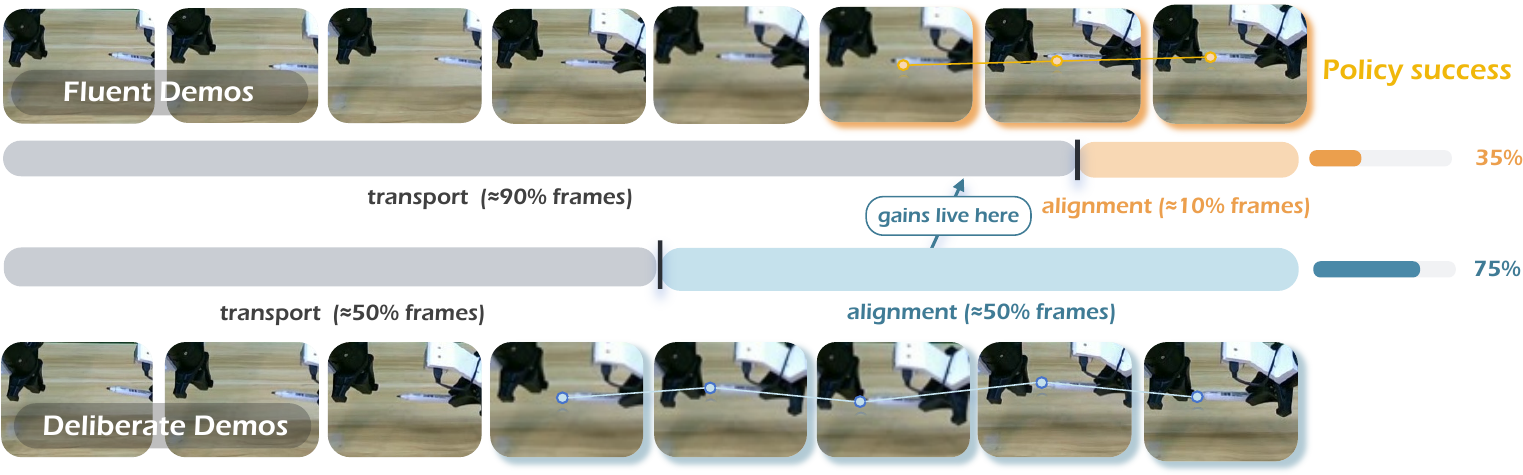} 
\end{center}
\vspace{-2mm}
\caption{\textbf{Where policy gains actually live.} On a pen-cap insertion task, a \emph{fluent} expert (top) spends $\approx\!90\%$ of frames on free-space transport and compresses the decisive alignment phase into $\approx\!10\%$ of frames, yielding $35\%$ success. A \emph{deliberate} demonstration (bottom) redistributes time toward alignment ($\approx\!50\%$ of frames), exposing the corrective micro-adjustments near contact and raising success to $75\%$. Success is decided in the alignment phase, yet fluent demonstrations supervise it the least.}
\label{fig:analysis}
\vspace{-2mm}
\end{figure}

The reason is intuitive yet easy to overlook: a demonstration optimized for human execution is not optimized for robot learning~\citep{laskey2017dart}. In most fine-manipulation trajectories, the vast majority of frames describe an easy free-space transport, while success is decided by a short alignment segment near contact. A fluent expert compresses this decisive segment into a handful of frames (Fig.~\ref{fig:analysis}, top: alignment occupies only $\approx\!10\%$ of frames, and the policy succeeds $35\%$ of the time). Because the imitation loss weights every frame equally, the policy receives abundant supervision for the easy transport and only faint supervision for the phase that actually determines success. Worse, an expert almost never enters a misaligned state, so the data shows the robot what success looks like but rarely how to recover toward it.

A first response is to fix the data, and we explore two remedies. Most directly, we change the demonstrations themselves: operators transport naturally but act as teachers near contact, slowing down, making small corrective motions, and occasionally approaching off-axis before recovering. We call these \emph{deliberate demonstrations} (Fig.~\ref{fig:analysis}, bottom). Alternatively, we keep the original data and resample it, upweighting frames around task-critical moments such as gripper changes, contact, and insertion~\citep{akgun2012keyframe,kou2024kisa,hejna2025robot,hejna2024re}. Both improve alignment success, confirming that the learning signal is imbalanced across the trajectory. Crucially, however, they help for different reasons: resampling merely replays the frames the dataset already contains, whereas deliberate collection expands the \emph{coverage} of correction and recovery states the policy would otherwise never see. This is why deliberate demonstrations win, and why data-only fixes hit a ceiling: slow collection is costly, and resampling can neither create recovery behavior nor add temporal detail that was never recorded.

These data-level fixes share a blind spot: they change how many alignment frames the policy sees, not how it sees each one. If the policy still maps a single static image to an action, the local motion that governs correction, whether the gripper is drifting, whether contact has begun, whether the error is shrinking, stays implicit. We therefore attack the bottleneck at the representation level and introduce \method, a compact dynamic feature for fine alignment: instead of reading a single frame, \method trains the vision-language model to predict how the local scene evolves over a short video neighborhood and uses this prediction to condition the action expert. The supervision is essentially free, distilled from short-horizon motion already recorded in every trajectory, so even fluent data yields dense signal about contact and correction. At deployment the neighborhood is no longer needed: the policy predicts the dynamic feature from the current frame alone, preserving the standard VLA interface. In short, \method does not ask for more frames, it extracts more signal from the frames already collected. Our contributions are threefold:
\begin{enumerate}[leftmargin=*, itemsep=0pt, topsep=2pt, parsep=0pt]
  \item We identify a counterintuitive failure mode of imitation learning: fluent expert demonstrations can hurt fine manipulation by compressing the critical alignment and recovery behaviors that decide success.
  \item We study two data-level remedies, deliberate slow-at-alignment collection and critical-segment resampling, and show that deliberate demonstrations win primarily by expanding recovery-state coverage rather than by reweighting existing frames.
  \item We introduce \method, a compact dynamic feature bridging the VLM and action expert that recovers most of the deliberate-demonstration gain from fluent data alone ($50.0$ to $62.2\%$ overall across six real-world tasks).
\end{enumerate}

%% file: sections/2_Related_Works.tex
\section{Related Work}
\label{sec:related}

\subsection{Vision-Language-Action Models}
Vision-language-action (VLA) models map language and visual observations to robot actions and are now a mainstream paradigm for general-purpose manipulation~\citep{kim2024openvla,zitkovich2023rt,wang2025vq, liu2025bridge, huang2025notvla, wang2026odyssey}, with recent work strengthening action parameterizations~\citep{intelligence2025pi_,black2024pi_0,bjorck2025gr00t} and latent representations~\citep{bu2025univla,cen2025worldvla,zhang2026dreamvla,liang2025mm,liu2025rdt, yang2026learning}. These models condition each action on a per-step visual state, yet fine alignment is decided by short critical segments that fluent demonstrations compress into few frames. \method keeps the VLA interface intact but replaces this static per-step state with a compact dynamic feature, conditioning the action expert on local motion rather than a single appearance snapshot.

\subsection{Robotic Representation Learning}
Robotic representation learning exposes task-relevant structure that raw images hide, via semantic or spatial priors~\citep{nair2022r3m,ma2022vip,caron2021emerging,oquab2023dinov2,karamcheti2023language, wog2026} or motion-oriented features~\citep{ye2025latent,chen2025moto,wang2025unified}. StaMo~\citep{stamo2025} differences two static state tokens, which breaks down when the decisive signal is short-horizon, while a complementary line predicts a compact future condition for general future modeling~\citep{wog2026}. \method shares the compact-condition view but targets the fine-alignment bottleneck: rather than differencing endpoints or summarizing the full horizon, it compresses a short temporal neighborhood around the action chunk into an action-shaped dynamic feature carrying the correction and contact cues fluent demonstrations leave implicit.

\subsection{Data Curation for Robot Learning}
Beyond scaling dataset coverage~\citep{o2024open,khazatsky2024droid,ebert2021bridge,chi2024universal}, data-centric work studies how composition affects transfer~\citep{yang2026data,hejna2025robot} and curates demonstrations by filtering for quality~\citep{kuhar2023learning,hejna2025robot}, rebalancing mixtures~\citep{hejna2024re}, retrieving relevant transitions~\citep{du2023behavior,lin2024flowretrieval}, or querying novel and risky states~\citep{cui2019uncertainty,hoque2021thriftydagger}. Our slow-at-alignment protocol is trajectory-centric, densifying supervision where visual change determines corrective action. Crucially, our controlled experiments separate two effects prior reweighting conflates: denser alignment frames help, but the larger gain comes from broader \emph{coverage} of correction and recovery states that resampling existing frames cannot provide. \method is the representation-side counterpart, recovering this signal from ordinary fluent demonstrations without curated collection.

%% file: sections/3_Method.tex
\section{Method}
\label{sec:method}
\subsection{Formalizing the Alignment Bottleneck}
\label{sec:method:bottleneck}
We begin by formalizing the bottleneck revealed by fluent demonstrations. A trajectory $\tau = \{(o_t, a_t)\}_{t=1}^{T}$ splits into a transport phase $\mathcal{T}_{\mathrm{tr}}$, where the robot moves toward the target, and an alignment phase $\mathcal{T}_{\mathrm{al}}$, where the final contact, insertion, or placement is decided. Let $\alpha = |\mathcal{T}_{\mathrm{al}}|/T$ be the fraction of frames spent in alignment. Since most imitation losses weight every step equally, the objective decomposes as
\begin{equation}
    \mathcal{L}_{\mathrm{BC}} = (1-\alpha)\,\mathcal{L}_{\mathrm{tr}} + \alpha\,\mathcal{L}_{\mathrm{al}},
    \label{eq:bc_phase_decomposition}
\end{equation}
so a fluent demonstration with small $\alpha$ spends almost all of its training weight on transport.

Yet importance does not scale with duration. With phase action-sensitivity $\sigma_\phi \propto \big|\partial P(\mathrm{success})/\partial a_t\big|$, fine manipulation obeys $\sigma_{\mathrm{al}} \gg \sigma_{\mathrm{tr}}$: a transport error is recoverable, while the same error near contact causes collision, jamming, or irreversible drift. Fluent demonstrations thus induce a \emph{duration--sensitivity mismatch}, allocating weight by how long a phase lasts rather than how much it decides success, and the signal is also temporal: a single frame near alignment rarely reveals whether to continue, stop, slide, or recover. We attack this bottleneck with three remedies across two levels: at the \emph{data level}, increasing the number of alignment frames (deliberate demonstrations) and their weight (critical-segment resampling); and at the \emph{representation level}, how each frame is encoded (dynamic visual features).

\subsection{Deliberate Demonstrations}
\label{sec:method:deliberate}

The most direct remedy is to change the demonstration protocol itself. Rather than executing the task as efficiently as possible, operators perform transport naturally but act like teachers near contact: they slow down, make small corrective motions, and occasionally approach with a mild offset before recovering. We call this \emph{deliberate demonstration}. By construction it raises the alignment-frame ratio, $\alpha_{\mathrm{delib}} > \alpha_{\mathrm{fluent}}$, so under the uniform loss of Eq.~\ref{eq:bc_phase_decomposition} the effective training weight on the high-sensitivity phase increases without touching the model or objective. More importantly, it exposes the local correction process that fluent experts compress away, how pose errors are reduced, how contact is handled, and how to recover from slight misalignment, precisely the corrective behaviors a policy needs at rollout. Deliberate demonstrations are thus not merely slower; they spend time where the policy is most sensitive. Sec.~\ref{sec:exp:main} compares fluent and deliberate data under a matched policy, task set, training budget, and trajectory count.

\subsection{Critical-Segment Resampling}
\label{sec:method:resampling}

Deliberate collection changes the data distribution but requires new demonstrations. As a training-time alternative, we keep the original trajectories fixed and instead sample task-critical segments more often. Given a set of key moments $\mathcal{K} = \{k_1,\ldots,k_M\}$ (gripper opening or closing, contact, insertion, placement), we form a critical window $\mathcal{W} = \{t : \min_{k\in\mathcal{K}} |t-k| \le r\}$ of radius $r$, upweight frames inside it by $\lambda$ ($w_t = \lambda$ if $t\in\mathcal{W}$, else $1$), and train with
\begin{equation}
    \mathcal{L}_{\mathrm{resample}}
    =
    \frac{1}{\sum_{t} w_t}
    \sum_{t=1}^{T}
    w_t\,\ell\bigl(\pi_\theta(o_t,l), a_t\bigr).
    \label{eq:resample_loss}
\end{equation}
Resampling is the training-time analogue of slowing down: it raises the effective contribution of alignment frames without altering the recorded trajectory, isolating how much of the gain is simply more supervision on high-sensitivity segments. Its limitation is equally clear: replaying the same frames cannot create missing recovery behavior or add temporal detail never captured, and over-weighting narrow windows reduces batch diversity. Sec.~\ref{sec:exp:main} compares resampling against deliberate demonstrations.

\subsection{Dynamic Alignment Features}
Deliberate collection and critical-segment resampling both expose more supervision around alignment, but the policy still has to infer local motion from isolated observations. \method addresses the same bottleneck at the representation level: it supplies the action expert with a compact dynamic feature for local motion during alignment, so correction and contact progress do not have to be recovered from a single image.

\label{sec:method:dynfeat}
\begin{wrapfigure}[13]{r}{0.32\textwidth}
    \vspace{-1.8\baselineskip}
    \centering
    \includegraphics[width=\linewidth]{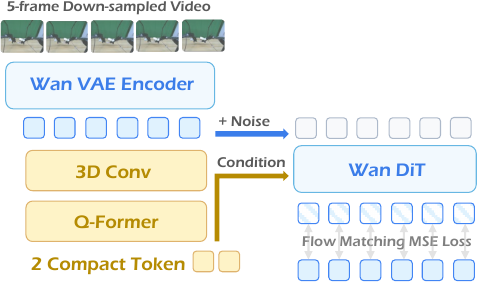}
    \caption{\textbf{Dynamic alignment feature.} }
    \label{fig:stair}
    \vspace{-0.8\baselineskip}
\end{wrapfigure}

For each training sample, we associate the supervised action chunk with a short dynamic observation clip,
\begin{equation}
    \mathcal{N}_t = \{o_{t+\delta}\}_{\delta \in \Delta},
\end{equation}
where $\Delta$ denotes a small set of temporal offsets around the action chunk. We first encode this future observation neighbourhood with a frozen WanVAE encoder $E_{\mathrm{wan}}$. A trainable projector $P_\psi$, implemented as a 3D convolution followed by a Q-Former, then compresses the video latent into compact dynamic tokens:
\begin{equation}
    \fdyn_t = P_\psi\!\left(E_{\mathrm{wan}}(\mathcal{N}_t)\right).
\end{equation}
The feature $\fdyn_t$ is not the full Wan VAE latent used by the flow-matching model. It is the compact token sequence produced after the frozen projector compresses the local video segment. Because the tokens are computed from local dynamics, they can carry cues such as moving into contact, slipping away, or correcting toward the target, while giving the policy a concentrated summary rather than raw frames or high-dimensional video latents.

The design keeps the compact-state view, but the token now summarizes local dynamics rather than static appearance. The policy does not need raw visual history if a small action-relevant token set preserves the changes needed for control. This is where fluent demonstrations are thinnest: the correction around contact is brief, but small differences there often decide success. The detailed training protocol is provided in the supplementary material.

\subsection{Training and Inference}
\label{sec:method:train}
\label{sec:method:training}

Our policy is a vision-language-action model that couples a VLM backbone~\citep{kim2024openvla} with a flow-matching DiT action expert. Given the current observation $o_t$ and language instruction $l$, the VLM produces a latent representation $z_t=f_{\mathrm{VLM}}(o_t,l)$. The action expert then predicts an action chunk $A_{t:t+H}$. The dynamic alignment feature is first mapped by an MLP into the action-conditioning space and then enters the action expert through cross-attention, so the action model conditions on both the current VLM state and local motion.

Training uses a two-stage protocol. In Stage I, only the WanVAE encoder $E_{\mathrm{wan}}$ is frozen. We jointly train the projector $P_\psi$, the MLP adapter, the action expert, and the selected VLA modules. Writing $\mathbf{c}^{\mathrm{dyn}}_t=\mathrm{MLP}_\eta(\fdyn_t)$ for the projected dynamic condition, the action flow-matching objective is
\begin{equation}
    \mathcal{L}_{\mathrm{I}}
    =
    \mathbb{E}_{s,A}
    \left[
    \left\|
    v_\theta(A_s,s,z_t,\mathbf{c}^{\mathrm{dyn}}_t)-v^\ast
    \right\|_2^2
    \right],
\end{equation}
where $s$ is the flow timestep and $v^\ast$ is the target velocity. The action gradient flows back through $\mathrm{MLP}_\eta$ and the projector $P_\psi$ (but not the frozen WanVAE encoder), shaping the dynamic tokens to be action-relevant.

In Stage II, the entire dynamic-feature pathway ($E_{\mathrm{wan}}$, the projector $P_\psi$, and the MLP adapter) is fixed to provide a stable target. \method adds a QFormer query module on top of the VLM hidden states and trains it to predict the same projected dynamic condition from the current observation and instruction:
\begin{equation}
    \widehat{\mathbf{c}}^{\mathrm{dyn}}_t = q_\phi\!\left(\mathrm{hidden}(f_{\mathrm{VLM}}(o_t,l))\right).
\end{equation}
We supervise this prediction with a cosine distance to the frozen projected-condition target,
\begin{equation}
    \mathcal{L}_{\mathrm{dyn}}
    =
    1-\mathcal{S}(\widehat{\mathbf{c}}^{\mathrm{dyn}}_t,\mathbf{c}^{\mathrm{dyn}}_t),
\end{equation}
and continue training the action expert with the frozen projected condition:
\begin{equation}
    \mathcal{L}_{\mathrm{II}}
    =
    \mathcal{L}_{\mathrm{action}}(z_t,\mathbf{c}^{\mathrm{dyn}}_t)
    +
    \lambda_{\mathrm{dyn}}\mathcal{L}_{\mathrm{dyn}}.
\end{equation}
During Stage II, the action expert still receives the frozen projected condition. The query module learns the target while the action condition stays fixed.

At inference time, the video neighbourhood is no longer available. The policy receives only the current RGB observation and instruction. The VLM produces $z_t$, the query module predicts $\widehat{\mathbf{c}}^{\mathrm{dyn}}_t$, and the action expert uses $(z_t,\widehat{\mathbf{c}}^{\mathrm{dyn}}_t)$ to generate the action chunk. Dynamic feature extraction is only used during training, while deployment keeps the same observation interface as a standard VLA policy.

%% file: sections/4_Exe.tex
\section{Experiments}
\label{sec:exp}

Our experiments are organized around four questions:
(Q1)~Do data-level remedies (deliberate collection and critical-segment resampling) improve fine alignment, and is the effect \emph{model-agnostic}? We test this with $\pi_{0.5}$ as an independent base (Sec.~\ref{sec:exp:datageneral}).
(Q2)~Can our model-level dynamic feature recover the deliberate-demonstration gain from fluent data alone on a controlled architecture (Sec.~\ref{sec:exp:main})?
(Q3)~Are the gains explained by recovery-state coverage, as the bottleneck analysis predicts (Sec.~\ref{sec:exp:coverage})?
(Q4)~Does \method remain competitive as a general policy (Sec.~\ref{sec:exp:simpler})?
Q1--Q3 are studied on a real robot across six tasks of increasing alignment difficulty; Q4 is verified on SimplerEnv.

\subsection{Setup}
\label{sec:exp:setup}

\begin{wrapfigure}{r}{0.2\textwidth}
    \centering
    \vspace{-10pt}
    \includegraphics[width=0.18\textwidth]{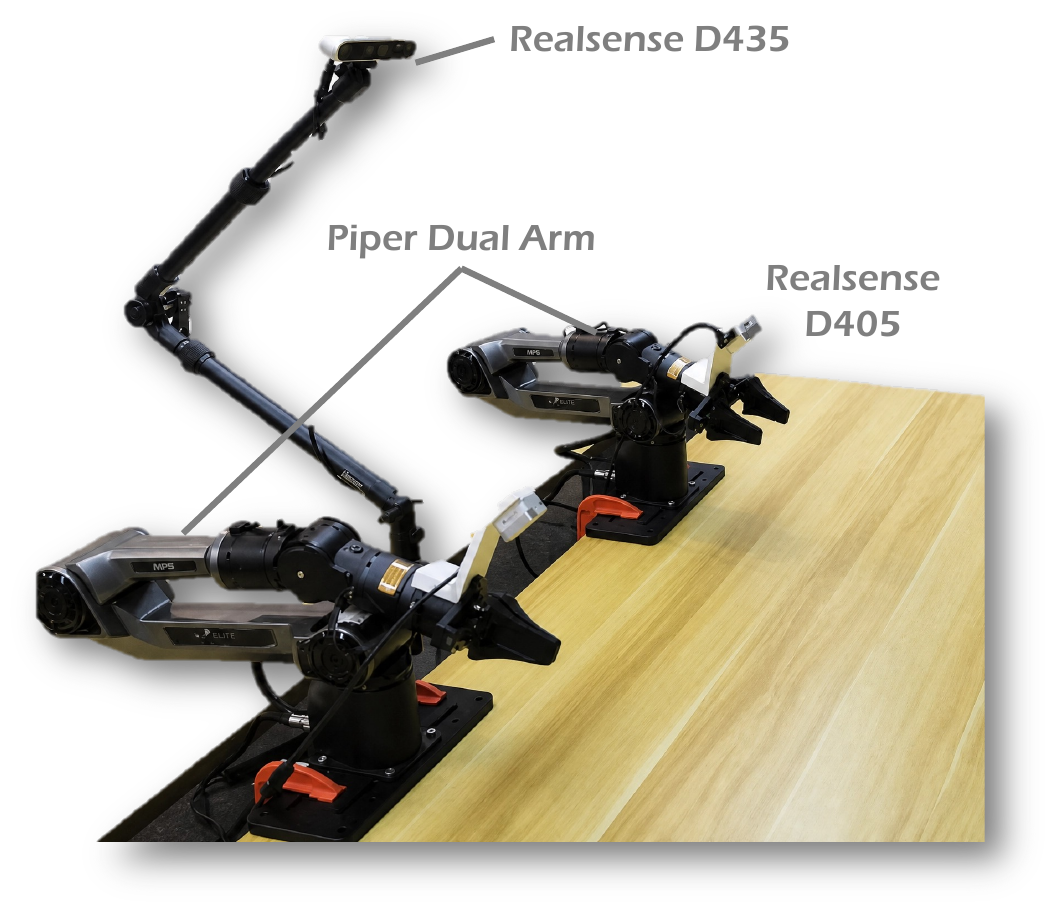}
    \caption{\textbf{Setup.}}
    \label{fig:real_robot_setup}
\end{wrapfigure}
\textbf{Hardware} All real-world experiments are conducted on a dual-arm Piper robot platform from AgileX Robotics, as shown in Fig.~\ref{fig:real_robot_setup}. For visual observation, we use one Intel RealSense D435 as the primary camera and two wrist-mounted Intel RealSense D405 cameras, one on each arm. All cameras operate at 30 FPS and capture images at a resolution of $640 \times 360$.

\begin{figure}[h]
\centering
\includegraphics[width=0.9\linewidth]{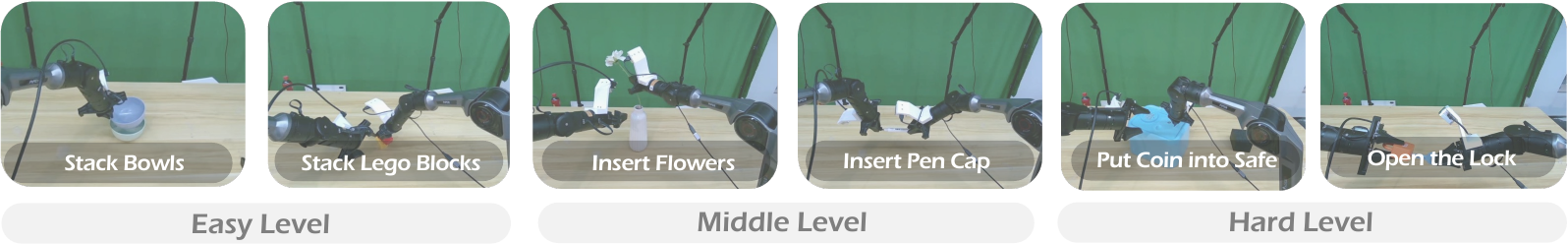}
\caption{\textbf{Six real-world manipulation tasks grouped by alignment difficulty.}}
\label{fig:real_tasks}
\end{figure}

\textbf{Tasks} We design six fine-grained manipulation tasks grouped into three difficulty levels by the alignment tolerance and the contact complexity of the final phase, as illustrated in Fig.~\ref{fig:real_tasks}. From easy to hard, the tasks are \emph{Stack Bowls}, \emph{Stack Lego Blocks}, \emph{Insert Flowers}, \emph{Insert Pen Cap}, \emph{Put Coin into Safe}, and \emph{Open the Lock}, with alignment requirements ranging from coarse position tolerance to multi-stage, contact-aware correction.

\textbf{Data conditions and protocol.} For every task we collect $100$ demonstrations and train by SFT on the mixture, evaluating each policy with $15$ trials per task. We compare three data conditions: \emph{Fluent} (efficient expert demonstrations), \emph{Deliberate} (slow-at-alignment demonstrations, matched at $100$ trajectories per task), and \emph{Resample} (critical-segment reweighting of the fluent set; Sec.~\ref{sec:method:resampling}). Fluent and deliberate sets are matched in trajectory count; deliberate trajectories are longer and thus contain more frames, an effect we isolate in Sec.~\ref{sec:exp:coverage}.

\textbf{Base policies.} To show the data-level effect is not tied to one architecture, we study it on two bases. \emph{Part~A} uses $\pi_{0.5}$~\citep{intelligence2025pi_}, a strong and independent policy, to establish that the effect is model-agnostic. \emph{Part~B} then studies our model-level remedy on a controlled VLA of our own (a VLM with a flow-matching DiT action expert; Sec.~\ref{sec:method:training}). Here the \emph{Baseline} is \method with the dynamic-feature pathway ablated: it keeps the same VLM, action expert, and query module, but is trained with the action loss only and without the video encoder, so that \emph{Baseline} and \method differ by exactly one factor, whether the conditioning tokens are supervised by the dynamic feature. Training schedules for both bases are given in the supplementary material.

\subsection{Data-Level Remedies are Model-Agnostic}
\label{sec:exp:datageneral}

We first establish, on the $\pi_{0.5}$ base, that the data-level remedies improve fine alignment independently of our model contribution. Table~\ref{tab:partA_pi05} reports success rates across the six tasks. Deliberate collection improves over fluent demonstrations, and the gap widens with alignment difficulty; critical-segment resampling recovers part of this gain without new data. Because $\pi_{0.5}$ shares none of our model-level components, this isolates the data effect itself.

\begin{table}[t]
\centering
\begin{minipage}[t]{0.40\textwidth}
\centering
\caption{\textbf{Part~A: $\pi_{0.5}$ base.} Success rate (\%), averaged by difficulty level. Data effect on an independent policy.}
\label{tab:partA_pi05}
\setlength{\tabcolsep}{4pt}
\resizebox{\linewidth}{!}{%
\begin{tabular}{l c c c c}
\toprule
Data & Easy & Mid. & Hard & Overall \\
\midrule
Fluent     & 90.0\% & 56.7\% & 10.0\% & 52.2\% \\
Deliberate & \textbf{93.3\%} & \textbf{73.3\%} & \textbf{36.7\%} & \textbf{67.8\%} \\
+ Resample & 90.0\% & 66.7\% & 20.0\% & 58.9\% \\
\bottomrule
\end{tabular}}
\end{minipage}
\hfill
\begin{minipage}[t]{0.55\textwidth}
\centering
\caption{\textbf{Part~B: Controlled study on our VLA base.}
}
\label{tab:main_level}
\setlength{\tabcolsep}{4pt}
\resizebox{\linewidth}{!}{%
\begin{tabular}{l l c c c c}
\toprule
Method & Data & Easy & Mid. & Hard & Overall \\
\midrule
Baseline         & Fluent      & 90.0\% & 40.0\% & 3.3\%  & 44.4\% \\
Baseline         & Deliberate  & 93.3\% & 56.7\% & 20.0\% & 56.7\% \\
+ Resample       & Fluent      & 90.0\% & 46.7\% & 10.0\% & 48.9\% \\
\rowcolor{gray!10}
+ \method (ours) & Fluent      & 93.3\% & 56.7\% & 13.3\% & 54.4\% \\
\rowcolor{gray!10}
+ \method (ours) & Deliberate  & \textbf{96.7\%} & \textbf{63.3\%} & \textbf{26.7\%} & \textbf{62.2\%} \\
\bottomrule
\end{tabular}}
\end{minipage}
\end{table}

\subsection{STAIR on a Controlled Architecture}
\label{sec:exp:main}

Table~\ref{tab:main_level} reports success rates on our VLA base, averaged within each difficulty level and overall. Three patterns emerge.
\emph{First,} the data trend of Part~A reappears on this architecture: the deliberate baseline beats the fluent one by a margin that widens with alignment difficulty, and resampling the fluent set recovers part of the gap, confirming the data effect is not specific to $\pi_{0.5}$.
\emph{Second,} \method trained on fluent data alone recovers most of the deliberate-demonstration gain, approaching the deliberate baseline without changing the data, which is the central result of the paper.
\emph{Third,} \method trained on deliberate data is strongest overall, showing that the data-level and model-level remedies are complementary.
Table~\ref{tab:main_per_task} reports per-task success rates, confirming the averaged trend is not driven by a single dominant task. The largest single-task gaps appear on \emph{Insert Pen Cap}, \emph{Put Coin into Safe}, and \emph{Open the Lock}, which require multi-millimeter alignment under partial occlusion.

\begin{table}[t]
\centering
\caption{\textbf{Per-task success rates (\%) on the six real-world tasks.} Same setup as Table~\ref{tab:main_level}, broken out per task.}
\label{tab:main_per_task}
\setlength{\tabcolsep}{4.5pt}
\resizebox{\linewidth}{!}{
\begin{tabular}{l l c c | c c | c c | c}
\toprule
& & \multicolumn{2}{c|}{Easy} & \multicolumn{2}{c|}{Middle} & \multicolumn{2}{c|}{Hard} & \\
Method & Data & Stack Bowls & Stack Lego & Insert Flowers & Insert Pen Cap & Coin into Safe & Open Lock & Avg \\
\midrule
Baseline           & Fluent      & 93.3\% & 86.7\% & 53.3\% & 26.7\% & 6.7\%  & 0.0\%  & 44.4\% \\
Baseline           & Deliberate  & 93.3\% & 93.3\% & 66.7\% & 46.7\% & 26.7\% & 13.3\% & 56.7\% \\
+ Resample         & Fluent      & 93.3\% & 86.7\% & 60.0\% & 33.3\% & 13.3\% & 6.7\%  & 48.9\% \\
\rowcolor{gray!10}
+ \method (ours)   & Fluent      & 93.3\% & 93.3\% & 66.7\% & 46.7\% & 20.0\% & 6.7\%  & 54.4\% \\
\rowcolor{gray!10}
+ \method (ours)   & Deliberate  & \textbf{100.0\%} & \textbf{93.3\%} & \textbf{73.3\%} & \textbf{53.3\%} & \textbf{33.3\%} & \textbf{20.0\%} & \textbf{62.2\%} \\
\bottomrule
\end{tabular}}
\end{table}

\subsection{Why Deliberate Demonstrations Outperform Resampling}
\label{sec:exp:coverage}

\begin{figure}[h]
\centering
\includegraphics[width=\linewidth]{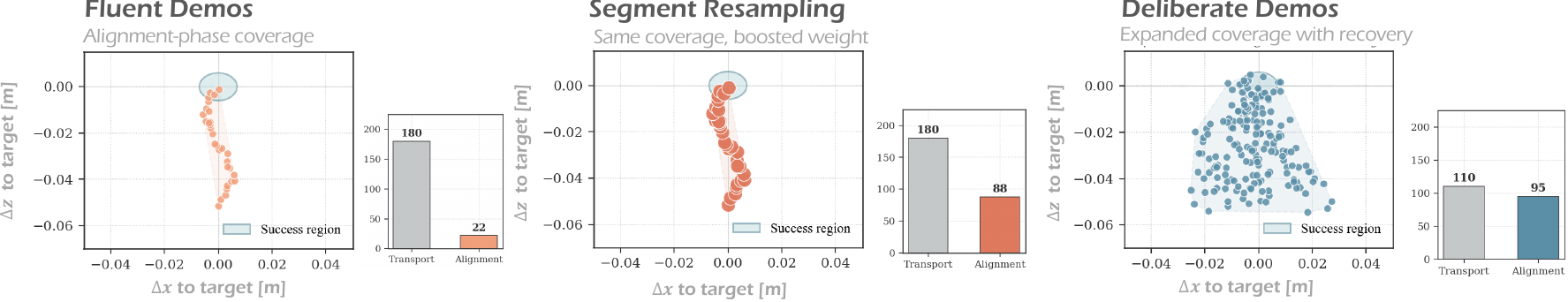}
\caption{
\textbf{Alignment-phase coverage vs.\ effective training weight.}
Each panel is one demonstration set. The scatter plots the end-effector position relative to the target $(\Delta x, \Delta z)$ over all alignment-phase frames, with the success region in blue; a wider spread means broader state coverage. The inset bar chart reports the \emph{average number of frames per trajectory} in the transport and alignment phases (for resampling, the alignment bar is the effective count after up-weighting). Resampling raises the alignment count from $22$ to $88$ yet keeps fluent's narrow coverage, whereas deliberate demonstrations reach a comparable count ($95$) while expanding coverage with off-axis approaches and recovery.
}
\label{fig:alignment-coverage}
\end{figure}

Deliberate demonstrations consistently outperform resampling (Table~\ref{tab:main_level}) even though both raise the effective alignment weight. Fig.~\ref{fig:alignment-coverage} explains why: we tune resampling so its alignment budget ($88$ frames per trajectory) matches deliberate's ($95$), holding the weight fixed, yet resampling inherits the narrow state coverage of the fluent set while deliberate substantially expands it with off-axis approaches and recovery. The remaining gap is therefore a coverage effect, not a weight effect: once alignment is sufficiently weighted, success is limited by whether the data exposes the local correction process, which replaying the same frames cannot create. This is precisely the gap that the model-level dynamic feature (Sec.~\ref{sec:method:dynfeat}) closes by extracting the correction signal from video.

\subsection{Generalization on SimplerEnv}
\label{sec:exp:simpler}

To test whether the dynamic feature is generally useful rather than specific to fine alignment, we evaluate \method on SimplerEnv~\citep{li2024evaluating}, a standard general-manipulation benchmark, against a broad set of state-of-the-art VLA models. Here \method is trained purely on large-scale robot data (no fluent/deliberate distinction): pretrained on the OXE mixture, then trained on Bridge and Fractal, and at inference it receives only the current RGB observation, exactly like a standard VLA. We report $30$ closed-loop rollouts on each of nine tasks (five Google Robot, four WidowX Bridge); training and protocol details are in the supplementary material.

As shown in Tables~\ref{tab:simplerenv} and~\ref{tab:simplerenv-widow}, \method is competitive with state-of-the-art generalist policies on both robots, with gains spanning grasping and final placement. Conditioning on compact dynamic feature thus benefits general manipulation broadly, not only the fine-alignment tasks that motivate it.

\begin{table*}[!t]
\centering

\caption{\textbf{SimplerEnv evaluation across different models on Google Robot tasks.}}
\label{tab:simplerenv}
\resizebox{0.7\textwidth}{!}{%
\begin{tabular}{l|cccc|cccc|c}
\toprule
\multirow{2}{*}{\textbf{Model}}
& \multicolumn{4}{c|}{\textbf{Visual Matching}}
& \multicolumn{4}{c|}{\textbf{Variant Aggregation}}
& \textbf{Overall} \\
& Pick Coke & Move Near & Drawer & Avg.
& Pick Coke & Move Near & Drawer & Avg.
& Avg. \\
\midrule
RT-1-X~\citep{brohan2022rt}
& 56.7\% & 31.7\% & 59.7\% & 53.4\%
& 49.0\% & 32.3\% & 29.4\% & 39.7\%
& 46.6\% \\

Octo-Base~\citep{team2024octo}
& 17.0\% & 4.2\% & 22.7\% & 16.8\%
& 0.6\% & 3.1\% & 1.1\% & 1.2\%
& 9.0\% \\

$\pi_0$~\citep{black2410pi0}
& 72.7\% & 65.3\% & 38.3\% & 58.8\%
& 75.2\% & 63.7\% & 25.6\% & 54.8\%
& 56.8\% \\

$\pi_0$-FAST~\citep{pertsch2025fast}
& 75.3\% & 67.5\% & 42.9\% & 61.9\%
& 77.6\% & 68.2\% & \textbf{31.3\%} & 59.0\%
& 60.5\% \\

OpenVLA~\citep{kim2024openvla}
& 16.3\% & 46.2\% & 35.6\% & 32.7\%
& 54.5\% & 47.7\% & 17.7\% & 39.8\%
& 33.8\% \\

GR00T-N1~\citep{bjorck2025gr00t}
& 47.0\% & 70.0\% & 18.1\% & 45.0\%
& 78.8\% & 62.5\% & 13.2\% & 51.5\%
& 48.4\% \\

Moto~\citep{chen2025moto}
& 74.0\% & 60.4\% & 43.1\% & 59.2\%
& -- & -- & -- & -- & -- \\

\rowcolor[HTML]{EFEFEF}
\textbf{Ours}
& \textbf{84.0\%} & \textbf{72.5\%} & \textbf{62.3\%}
& \textbf{72.9\%}
& \textbf{81.9\%} & \textbf{75.0\%} & 27.3\%
& \textbf{61.4\%} & \textbf{67.2\%} \\
\bottomrule
\end{tabular}%
}

\vspace{8pt}

\caption{\textbf{SimplerEnv evaluation across different models on WidowX Robot tasks.}}
\label{tab:simplerenv-widow}
\resizebox{\textwidth}{!}{%
\begin{tabular}{l|cc|cc|cc|cc|cc}
\toprule
\multirow{2}{*}{\textbf{Model}}
& \multicolumn{2}{c|}{\textbf{Put Spoon on Towel}}
& \multicolumn{2}{c|}{\textbf{Stack Green on Yellow}}
& \multicolumn{2}{c|}{\textbf{Put Carrot on Plate}}
& \multicolumn{2}{c|}{\textbf{Put Eggplant in Basket}}
& \multicolumn{2}{c}{\textbf{Overall Average}} \\
& Grasp Spoon & Success
& Grasp G Block & Success
& Grasp Carrot & Success
& Grasp Eggplant & Success
& Grasp Avg. & Success Avg. \\
\midrule
RT-1-X~\citep{brohan2022rt}
& 16.7\% & 0.0\%
& 8.3\% & 0.0\%
& 20.8\% & 4.2\%
& 0.0\% & 0.0\%
& 11.5\% & 1.1\% \\

Octo-Base~\citep{team2024octo}
& 34.7\% & 12.5\%
& 31.9\% & 0.0\%
& 52.8\% & 8.3\%
& 66.7\% & 43.1\%
& 46.5\% & 16.0\% \\

SpatialVLA~\citep{wu2025spatial}
& 25.0\% & 20.8\%
& 58.3\% & 25.0\%
& 41.7\% & 20.8\%
& 79.2\% & 70.8\%
& 51.1\% & 34.4\% \\

$\pi_0$~\citep{black2410pi0}
& 45.8\% & 29.1\%
& 25.0\% & 0.0\%
& 50.0\% & 16.7\%
& 91.6\% & 62.5\%
& 40.1\% & 27.1\% \\

$\pi_0$-FAST~\citep{pertsch2025fast}
& 62.5\% & 29.1\%
& 58.5\% & 21.9\%
& 54.0\% & 10.8\%
& 83.3\% & 66.6\%
& 48.3\% & 32.1\% \\

OpenVLA~\citep{kim2024openvla}
& 4.1\% & 0.0\%
& 33.0\% & 0.0\%
& 12.5\% & 0.0\%
& 8.3\% & 4.1\%
& 7.8\% & 1.1\% \\

GR00T-N1~\citep{bjorck2025gr00t}
& 83.3\% & 62.5\%
& 54.2\% & 45.8\%
& 70.8\% & 16.7\%
& 41.7\% & 20.8\%
& 49.5\% & 36.5\% \\

UniVLA~\citep{bu2025univla}
& 76.4\% & 52.8\%
& 66.7\% & 2.8\%
& \textbf{79.2\%} & 55.6\%
& 87.5\% & 66.7\%
& 77.5\% & 45.6\% \\

\rowcolor[HTML]{EFEFEF}
\textbf{Ours}
& \textbf{81.8\%} & \textbf{69.2\%}
& \textbf{73.1\%} & \textbf{62.3\%}
& 76.8\% & 57.4\%
& \textbf{88.6\%} & \textbf{73.4\%}
& \textbf{80.1\%} & \textbf{65.6\%} \\
\bottomrule
\end{tabular}%
}

\end{table*}

%% file: sections/5_Conclusion.tex
\section{Conclusion}
\label{sec:conclusion}

We studied why fluent expert demonstrations can be poor teachers for fine-grained manipulation: skilled operators compress the alignment and recovery motions that decide success into a brief window, leaving the policy under-supervised exactly where precision matters. We addressed this at two levels: deliberate collection and critical-segment resampling add supervision on these high-sensitivity moments, while \method distills the same signal into a compact dynamic feature shared by the VLM and the action expert. More broadly, our results argue that robust alignment depends less on perfectly efficient trajectories than on demonstrations and representations that make correction visible to the learner.

%% file: sections/6_limitation.tex
\section{Limitations}
\label{sec:limitation}

Our approach has several limitations. At inference time, the dynamic feature is predicted deterministically from the current observation, which cannot capture multi-modal or partially observed contact futures. It also relies primarily on visual dynamics, and therefore does not replace explicit geometric, tactile, or force feedback in highly constrained contact tasks. Finally, key-frame and neighbourhood selection are currently treated as hyperparameters; learning task-conditioned temporal selection remains an important direction for future work.